\documentclass[conference]{IEEEtran}

\usepackage{cite}
\usepackage{amsmath,amssymb,amsfonts}
\usepackage{algorithmic}
\usepackage{algorithm}
\usepackage{multirow}
\usepackage{graphicx}
\usepackage{textcomp}
\usepackage{xcolor, soul}
\usepackage{booktabs}
\usepackage{times}

\def\BibTeX{{\rm B\kern-.05em{\sc i\kern-.025em b}\kern-.08em
    T\kern-.1667em\lower.7ex\hbox{E}\kern-.125emX}}

\begin{document}

\title{Do Spikes Protect Privacy? \\ Investigating Black-Box Model Inversion Attacks \\ in Spiking Neural Networks}

\author{\IEEEauthorblockN{Hamed Poursiami}
\IEEEauthorblockA{\textit{ECE Department} \\
George Mason University \\ Fairfax, VA, USA\\
Email: hpoursia@gmu.edu}

\and
\IEEEauthorblockN{Ayana Moshruba}
\IEEEauthorblockA{\textit{ECE Department} \\
George Mason University \\ Fairfax, VA, USA\\
Email: amoshrub@gmu.edu}

\and
\IEEEauthorblockN{Maryam Parsa}
\IEEEauthorblockA{\textit{ECE Department} \\
George Mason University \\ Fairfax, VA, USA\\
Email: mparsa@gmu.edu}

}

\maketitle
\begin{abstract}

As machine learning models become integral to security-sensitive applications, concerns over data leakage from adversarial attacks continue to rise. Model Inversion (MI) attacks pose a significant privacy threat by enabling adversaries to reconstruct training data from model outputs. While MI attacks on Artificial Neural Networks (ANNs) have been widely studied, Spiking Neural Networks (SNNs) remain largely unexplored in this context. Due to their event-driven and discrete computations, SNNs introduce fundamental differences in information processing that may offer inherent resistance to such attacks. A critical yet underexplored aspect of this threat lies in black-box settings, where attackers operate through queries without direct access to model parameters or gradients, representing a more realistic adversarial scenario in deployed systems. This work presents the first study of black-box MI attacks on SNNs. We adapt a generative adversarial MI framework to the spiking domain by incorporating rate-based encoding for input transformation and decoding mechanisms for output interpretation. Our results show that SNNs exhibit significantly greater resistance to MI attacks than ANNs, as demonstrated by degraded reconstructions, increased instability in attack convergence, and overall reduced attack effectiveness across multiple evaluation metrics. Further analysis suggests that the discrete and temporally distributed nature of SNN decision boundaries disrupts surrogate modeling, limiting the attacker’s ability to approximate the target model.

\end{abstract}

\begin{IEEEkeywords}
Spiking Neural Networks, Black-Box Model Inversion Attacks, Neuromorphic Privacy, Adversarial Machine Learning
\end{IEEEkeywords}

\section{Introduction}
\label{sec:intro}

\noindent
Spiking Neural Networks (SNNs) have emerged as a promising computational paradigm, inspired by the biological processes of neuronal communication. By processing information through discrete spike events, SNNs offer a more biologically plausible and efficient alternative to traditional Artificial Neural Networks (ANNs) \cite{nunes2022spiking}. Their event-driven nature, in which neurons remain mostly inactive until a spike occurs, enables SNNs to excel in applications such as real-time object detection \cite{cordone2022object, barchid2021deep}, speech recognition\cite{wu2020deep}, and robotics \cite{bing2018survey}, where energy efficiency and temporal processing are crucial~\cite{poursiami2025scalable}. However, as their use expands into critical areas, understanding and addressing their potential privacy risks is essential to maintain the integrity and confidentiality of sensitive information\cite{kim2022privatesnn,poursiami2024watermarking,tahmasivand2025lm,moshruba2025neuromorphic}.

Among the various privacy threats, Model Inversion (MI) attacks stand out due to their potential to reconstruct sensitive training data from a model's outputs. These attacks exploit the learned patterns of the model to reverse-engineer sensitive information, often revealing training data or inferring private attributes \cite{fredrikson2015model}. MI attacks can be divided into two main categories: white-box and black-box. In white-box scenarios, attackers have full access to the model's architecture and parameters, often employing gradient-based optimization for precise reconstructions. In contrast, black-box attacks operate under more restrictive conditions, relying solely on input-output queries to infer private information  \cite{dibbo2023sok}. Given the real-world relevance of black-box scenarios, their impact on SNN privacy warrants deeper investigation \cite{dionysiou2023exploring}.

Recent advancements in black-box MI attacks have introduced diverse methodologies to extract sensitive information from machine learning models. A prominent approach involves generative models, such as Generative Adversarial Networks (GANs) and Variational Autoencoders (VAEs), which optimize in a latent space rather than directly in the input space \cite{nguyen2024label,han2023reinforcement, kahla2022label}. These models iteratively adjust their latent variables to reconstruct candidate inputs that align with the target model’s outputs. Some methods further enhance the reconstruction quality by leveraging a learned prior from auxiliary public datasets for pre-training or fine-tuning \cite{khosravy2022model, xu2023sparse,ye2023c2fmi}. While effective, this reliance on auxiliary datasets limits their applicability when such data are unavailable or do not align with the target domain. To address this, agnostic black-box methods have emerged, which operate independently of external datasets or assumptions about the target data distribution \cite{tramer2016stealing}. One such approach, GAMIN \cite{aivodji2019gamin}, combines adversarial optimization and surrogate modeling to infer sensitive inputs solely through direct queries to the target model.

In SNNs, however, the prior studies have primarily explored privacy-preserving mechanisms and adversarial robustness. For example, differential privacy has been integrated into SNNs to protect training data by introducing noise during gradient updates, as demonstrated by \cite{wang2022dpsnn}. Similarly, \cite{kim2022privatesnn} addresses data and class leakage during ANN-to-SNN conversion through synthetic data generation and temporal learning rules. Other efforts, such as \cite{moshruba2025neuromorphic,moshruba2025privacy, moshruba2025izhikevich}, have highlighted the relative resilience of SNNs against membership inference attacks due to their non-differentiable spiking dynamics. In the context of MI attacks, BrainLeaks \cite{poursiami2024brainleaks} developed a tailored gradient-based attack to overcome the challenges of SNNs' spiking mechanisms, demonstrating that while these models exhibit some resistance, they remain vulnerable to such privacy breaches. However, the privacy characteristics of SNNs under black-box MI attacks remain unexplored, leaving a significant gap in understanding their behavior in restricted-access scenarios.

This gap highlights the need to systematically evaluate whether SNNs' spiking mechanisms confer unique privacy-preserving properties or render them susceptible to black-box MI attacks. To address this challenge, this paper investigates the behavior of SNNs under black-box MI attacks using the GAMIN framework. By leveraging GAMIN’s agnostic black-box setting, we provide a detailed evaluation of how SNNs respond to these attacks compared to ANNs. The primary contributions of this work are as follows:
\begin{enumerate}
    \item Conducting the first comprehensive evaluation of SNNs under black-box MI attacks through extensive experiments.
    \item Analyzing and comparing the susceptibility of SNNs and ANNs to black-box MI attacks, with a focus on evaluating differences arising from their computational paradigms.
    \item Offering new insights into the interplay between spiking dynamics and privacy, advancing our understanding of SNN-specific privacy characteristics.
\end{enumerate}

By addressing this unexplored domain, this work establishes a foundation for future research in the privacy of SNNs and neuromorphic computing.

\section{Preliminary}
\label{sec:preliminary}

\subsection{Spiking Neural Networks (SNNs)}

\noindent
SNNs are biologically inspired computational models designed to emulate the temporal dynamics and communication mechanisms of biological neurons. Unlike ANNs, which rely on continuous-valued activations, SNNs process information via discrete, time-dependent spikes \cite{schuman2022opportunities}. Each neuron in an SNN accumulates input spikes over time. When the membrane potential crosses a threshold, the neuron generates an output spike that propagates to downstream neurons. This event-driven communication makes SNNs energy-efficient and well-suited for neuromorphic applications.

Among various spiking neuron models, the \textit{Leaky Integrate-and-Fire (LIF)} model is widely adopted due to its simplicity and biological relevance \cite{izhikevich2004model}. The membrane potential dynamics of a single LIF neuron in discrete time can be expressed as:

\begin{equation}
    \nu[n] = \alpha \cdot \nu[n-1] + \sum_{k} \omega_k \cdot I_k[n] - O[n-1] \cdot \eta \ , 
\end{equation}
where $\nu[n]$ represents the membrane potential at time step $n$, $\alpha$ is the leakage factor that models the decay of potential over time, $I_k[n]$ denotes the spike input from presynaptic neuron $k$, and $\omega_k$ is the corresponding synaptic weight. The neuron’s output spikes, $O[n]$, are determined using a thresholding function:

\begin{equation}
    O[n] = 
    \begin{cases} 
        1, & \text{if } \nu[n] > \eta \\ 
        0, & \text{otherwise}.
    \end{cases}
    \label{eq:thresholding}
\end{equation}
When the membrane potential exceeds the firing threshold $\eta$, the neuron generates a spike and undergoes a soft-reset mechanism, subtracting the threshold from the potential to avoid continuous firing.

\begin{figure}[t!]
  \centering
  \includegraphics[width=\linewidth]{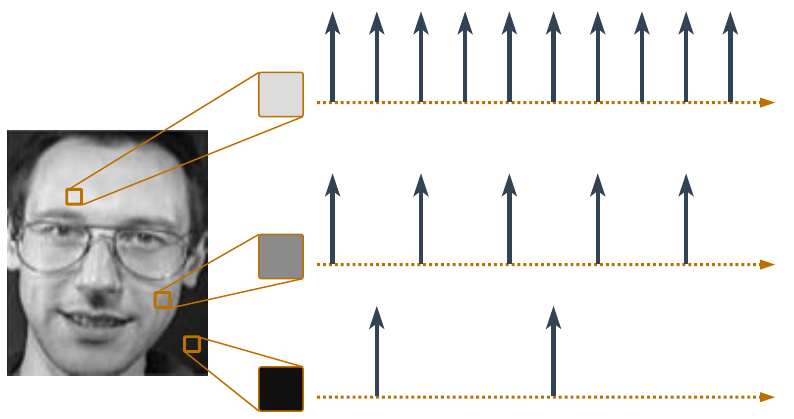}
  \caption{Illustration of rate encoding: Higher pixel intensities correspond to higher spike rates, while lower intensities produce sparser spike trains.}
  \label{fig:rate_encoding}
\end{figure}

\begin{figure*}[htbp!]
  \centering
  \includegraphics[width=\textwidth]{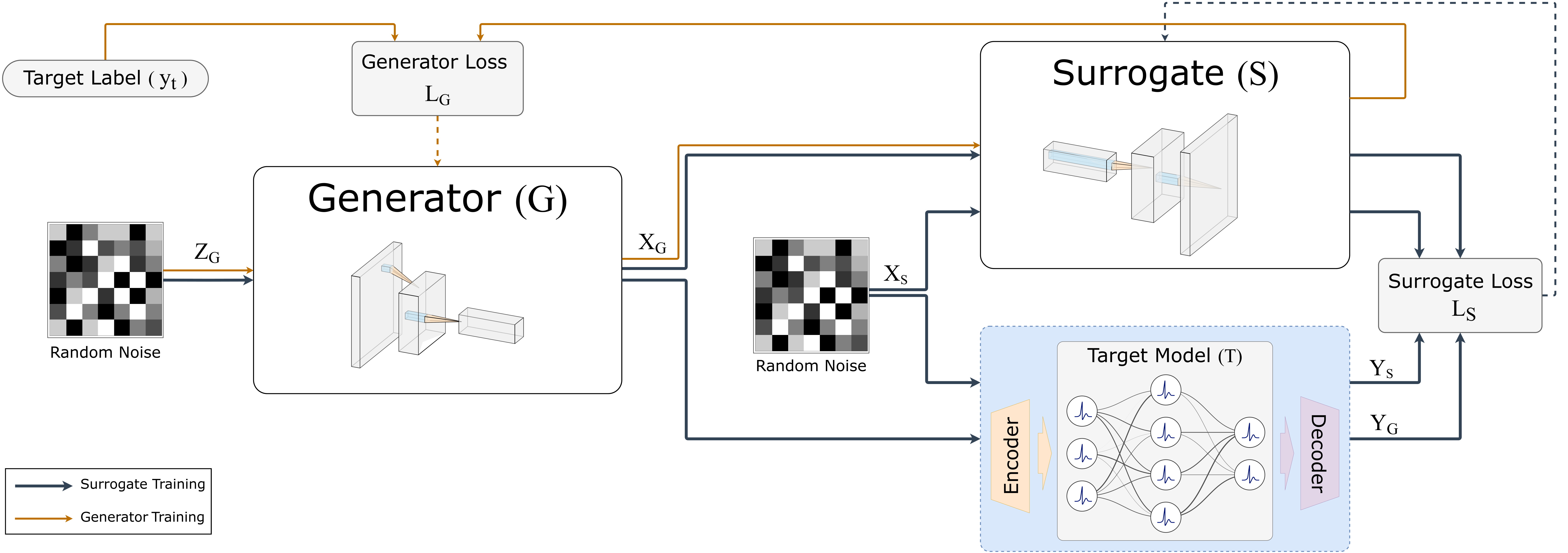}
  \caption{Workflow of the GAMIN framework applied to an SNN target model. The surrogate model and generator are trained iteratively, with dark blue arrows representing the surrogate training phase and orange arrows indicating the generator training phase.}
  \label{fig:Gamin_SNN}
\end{figure*}

For event-driven systems, input spikes can be directly obtained from neuromorphic sensors such as \textit{Dynamic Vision Sensors (DVS)}, which naturally generate spikes in response to changes in visual stimuli \cite{hu2016dvs}. These sensors provide asynchronous spike-based data, making them inherently compatible with SNNs without requiring additional encoding.

However, to process non-spiking data in SNNs, inputs must be converted into spike trains using an encoding mechanism. One commonly used approach is \textit{rate encoding}, where the spike frequency represents input intensity: higher input values correspond to higher spike rates, while lower values result in sparser spikes (see Figure~\ref{fig:rate_encoding}).  In this scheme, a continuous input value is mapped to a spike train by generating spikes with a probability proportional to the input magnitude over a fixed time window \cite{gerstner2014neuronal}.

Training SNNs with backpropagation-based learning algorithms poses unique challenges due to the non-differentiable nature of spike-based activity. Particularly, the discontinuity of the spike generation function in Equation~\ref{eq:thresholding} blocks gradient flow, making standard backpropagation ineffective. To address this, one approach is to use \textit{surrogate gradient methods}. These methods approximate the spiking function with a smooth, differentiable alternative during the backward pass, enabling techniques like \textit{Backpropagation Through Time (BPTT)} to be applied effectively \cite{bellec2018long, neftci2019surrogate}.


\subsection{Model Inversion (MI) Attacks}

\noindent
MI attacks aim to extract sensitive information from machine learning models by leveraging their learned representations. Formally, given a target model $T: \mathbb{R}^d \to \mathbb{R}^k$ and an observed output $\mathbf{y} = T(\mathbf{X})$ the objective of an MI attack is to approximate the input  $\mathbf{X} \in \mathbb{R}^d$ that produced the output $\mathbf{y} \in \mathbb{R}^k$. This is often framed as an optimization problem \cite{struppek2022plug}:
\begin{equation}
\hat{\mathbf{X}} = \arg\min_{\mathbf{X} \in \mathbb{R}^d} \mathcal{L}(T(\mathbf{X}), \mathbf{y}),
\label{eq:MI-opt}
\end{equation}
where $\mathcal{L}(\cdot, \cdot)$ is a loss function measuring the similarity between the model’s output for a candidate input $\mathbf{X}$ and the target output $\mathbf{y}$. In white-box scenarios, attackers often leverage gradient information from the model to iteratively solve this optimization problem. Black-box scenarios, on the other hand, are more challenging, as they rely solely on input-output queries without access to internal parameters or gradients. 
One popular approach to tackle black-box MI attacks is to train a surrogate model that approximates the behavior of the target model. By querying the target model with synthetic inputs and observing its responses, the surrogate model learns to mimic the target model’s decision boundaries, enabling attackers to indirectly reconstruct input  data\cite{fang2024privacy}.

\section{Black-Box Model Inversion on SNNs}
\label{sec: methodology}

\noindent
In this work, we utilize GAMIN (Generative Adversarial Model INversion) \cite{aivodji2019gamin}, a framework originally proposed for black-box model inversion on ANNs, and adapt it for the SNN domain. GAMIN works by training two neural networks in tandem: \textbf{a generator} $G$, which maps random noise \mbox{$Z_G \sim \mathcal{N}(0, 1)$} to synthetic inputs $X_G$, and \textbf{a surrogate model} $S$, which approximates the target model $T$. The goal is to generate inputs that, when passed through the surrogate, match the desired target label $y_t$ output by $T$. The training process consists of the following steps:

\begin{enumerate}
    \item The generator $G$ produces synthetic inputs $X_G = G(Z_G)$ from random noise.
    \item The target model $T$ is queried using both $X_G$ and additional random inputs $X_S$, generating predictions $Y_G = T(X_G)$ and $Y_S = T(X_S)$.
    \item The surrogate model $S$ is trained to mimic the target model $T$ while distinguishing noise inputs $X_S$ from generated samples $X_G$. This is achieved using a boundary-equilibrium loss:
    \begin{equation}
        L_S = L_H(X_S, Y_S) - k_t \cdot L_H(X_G, Y_G),
    \end{equation}
    
    where $L_H$ is the cross-entropy loss, and $k_t$ is a dynamically adjusted equilibrium factor updated as:
    \begin{equation}
        k_{t+1} = k_t + \lambda_k (\gamma_k L_H(X_S, Y_S) - L_H(X_G, Y_G)),
    \end{equation}
    
    Here, $\lambda_k$ and $\gamma_k$ are hyperparameters controlling the adjustment of $k_t$.

    \item The generator $G$ is optimized by minimizing the cross-entropy loss between the predictions of the combined model $S \circ G$ and the target label $y_t$:
    \begin{equation}
            L_G = L_H(S(G(Z_G)), y_t).
    \end{equation}

    During this step, the parameters of $S$ are kept fixed, ensuring that $G$ learns to generate inputs that are classified as $y_t$ by the surrogate model.

\end{enumerate}

\noindent
These steps are repeated iteratively until the generator produces high-quality inputs corresponding to the target label $y_t$, while the surrogate model closely approximates the decision boundaries of $T$.

To adapt GAMIN for SNNs, encoding and decoding mechanisms are utilized to handle the distinct input-output representations of these models. \textit{Rate encoding} transforms static input data, such as pixel intensities, into spike trains by mapping intensities to spike rates over a fixed temporal window  \cite{auge2021survey}. 

For output processing, \textit{decoding mechanisms} convert spiking activity into confidence scores that GAMIN can process. Common decoding methods include spike count, time-averaged firing rate, and membrane potential \cite{eshraghian2023training}.

It should be noted that encoding and decoding mechanisms are embedded components of the SNN model and do not interfere with GAMIN’s model-agnostic assumption. As a result, GAMIN treats SNNs and ANNs equivalently as queryable black-box models, operating independently of their underlying architecture or computational principles.

The application of GAMIN to SNNs, combined with its query-based agnostic framework, enables a direct investigation of the privacy vulnerabilities in spiking models. Figure \ref{fig:Gamin_SNN} illustrates the overall workflow, showing the interaction between the generator, surrogate model, and target model.

\section{Experiments}
\label{sec:experiments}

\subsection{Experimental Setup}

\noindent
We evaluate the effectiveness of the attack on two tasks: face recognition and digit classification. For face recognition, we use the AT\&T Face Database \cite{samaria1994orl}, which contains 400 grayscale images across 40 unique subjects. For digit classification, the MNIST \cite{deng2012mnist} dataset is utilized, containing 70,000 grayscale images of handwritten digits. For SNNs, static image data from both datasets are converted into 25-step spiking representations using a rate-encoding scheme, where pixel intensities are mapped to spike rates.

The target models, both ANNs and SNNs, share the same fully connected architecture with a single hidden layer of 3,000 neurons to allow fair comparisons. For SNNs, we employ LIF neurons with a leakage rate of $0.7$. The SNNs are trained using backpropagation through time (BPTT) \cite{bellec2018long} with a fast-sigmoid function (slope of 40) for surrogate gradient calculations \cite{neftci2019surrogate}. The loss function accumulates the softmax cross-entropy across all time steps, computed from the membrane potentials of the output neurons. Table~\ref{tab:target_accuracy} summarizes the test accuracy of the target models (ANNs and SNNs) trained on the MNIST and AT\&T Face datasets. 

We adopt the attack protocol from the GAMIN framework, which retains the best-performing model during training to prevent performance degradation from suboptimal updates, and reduce the impact of catastrophic forgetting.
The query budget is set to 1,280,000 queries (or 20,000 batches of 64), aligning with the original framework to provide adequate query capacity for comprehensive evaluation. All implementations were performed using PyTorch \cite{paszke2019pytorch} framework, with SNNtorch \cite{eshraghian2023training} employed for building and training the SNNs.

\subsection{Evaluation Metrics}

We employ five metrics to analyze different aspects of the black-box MI attack and evaluate the performance of the generator, and surrogate models.
Global Convergence Score (\(M_{\text{global}}\)), Fidelity Score (\(F_S\)), and Combined Accuracy (\(A_{S \circ G}\)) are adopted from GAMIN \cite{aivodji2019gamin}, while Surrogate Test Accuracy (\(A_S\)) and Target Model Accuracy on Inverted Samples (\(A_T\)) are introduced to provide additional insights into the surrogate and reconstructed samples.

\subsubsection{Global Convergence Score (\(M_{\text{global}}\))}
Inspired by the equilibrium loss in BEGAN \cite{berthelot2017began}, this metric assesses the overall performance of the attack by combining the surrogate model's fidelity and the generator’s effectiveness. A lower \(M_{\text{global}}\) value reflects improved alignment between the surrogate and generator and is defined as:
\begin{equation}
    M_{\text{global}} = L_H(X_S, Y_S) - |\lambda_k L_H(X_S, Y_S) - L_H(X_G, Y_G)|,
\end{equation}
where \(L_H(X_S, Y_S)\) is the loss of the surrogate model on random inputs, \(L_H(X_G, Y_G)\) is the loss of the surrogate on inputs generated by the generator, and \(\lambda_k\) is the dynamically adjusted equilibrium factor. \(M_{\text{global}}\) is also used during training to retain the best-performing generator-surrogate pair.

\begin{table}[tbp]
\centering
\caption{Accuracy of the target models on the test data}
\label{tab:target_accuracy}
\resizebox{0.8\columnwidth}{!}{%
\begin{tabular}{@{}ccc@{}}
\toprule
\textbf{Dataset}                     & \textbf{Model Type} & \textbf{Test Accuracy} \\ \midrule
\multirow{2}{*}{MNIST}      & ANN        & 97.71\%       \\
                            & SNN        & 97.67\%       \\ \midrule
\multirow{2}{*}{AT\&T Face} & ANN        & 95.00\%       \\
                            & SNN        & 91.25\%       \\ \bottomrule
\end{tabular}%
}
\end{table}

\begin{table*}[ht]
\centering
\renewcommand{\arraystretch}{1.1}
\caption{Evaluation Metrics for GAMIN Black-Box Model Inversion Attack on ANNs and SNNs}
\label{tab:evaluation_metrics}
\resizebox{0.8\textwidth}{!}{%
\begin{tabular}{@{}ccccccc@{}}
\toprule
Dataset                     & Model Type & $M_{global}$        & $F_S$  &  $A_S $    & $A_{S \circ G}$  & $A_T$         \\  \midrule
\multirow{2}{*}{MNIST}      & ANN        & 1.0371 $\pm$ 0.12& 0.8929                 & 68.13   $\pm$ 6.9               & 100           & 100           \\
                            & SNN        & 1.4904  $\pm$ 0.58               & \textbf{0.9541}        & 19.88  $\pm$ 4.7                & 70.0          & 70.0          \\ \midrule
\multirow{2}{*}{AT\&T Face} & ANN        & 3.4326 $\pm$ 0.33               & 0.9868                 & 29.32   $\pm$ 6.6               & 92.5          & 97.5          \\
                            & SNN        & 3.9378  $\pm$ 0.65              & 0.9948                 & \textbf{ 6.18} $\pm$ 4.5         & \textbf{37.5} & \textbf{49.2} \\ \bottomrule
\end{tabular}%
}
\end{table*}

\subsubsection{Surrogate Fidelity (\(F_S\))} This metric evaluates how well the surrogate model approximates the target model’s behavior. It is computed as:
\begin{equation}
    F_S = 1 - \text{MAE}(y, \hat{y}),
\end{equation}
where \(y\) and \(\hat{y}\) are the predictions of the target model and surrogate model, respectively, on a batch of random inputs. A higher \(F_S\) value indicates better fidelity and corresponds to a closer alignment with the target model’s decision boundaries.

\subsubsection{Surrogate Test Accuracy (\(A_S\))} To assess the surrogate model’s generalization ability, we compute its classification accuracy on the original test set. This complements \(F_S\) by evaluating how well the surrogate performs on real data, as opposed to random or generated inputs.

\subsubsection{Combined Accuracy (\(A_{S \circ G}\))}
This metric evaluates the ability of generator \(G\)  to synthesize inputs that the surrogate \(S\) model classifies as belonging to the target class. It is derived by computing the categorical accuracy of the surrogate on the generator’s outputs, representing the proportion of generated samples assigned to the intended label.

\subsubsection{Target Model Accuracy on Inverted Samples (\(A_T\))} This metric evaluates the quality of the reconstructed inputs by passing a batch of inverted samples through the target model \(T\). It measures how accurately the target model classifies the generator’s outputs, providing direct insight into the effectiveness of the inversion attack.

\section{Discussion}
\label{sec:discussion}

\noindent
Table~\ref{tab:evaluation_metrics} presents the evaluation metrics for GAMIN black-box MI attacks on ANNs and SNNs, highlighting key differences in attack performance across both architectures.

The M-global scores, which reflect overall attack performance by providing insight into the convergence of the surrogate and generator networks, are consistently higher for SNNs. Additionally, the combined accuracy \( A_{S \circ G} \), which measures how well the generator aligns its outputs with the surrogate, drops substantially for SNNs, indicating that the attack struggles to converge and generate high-quality reconstructions in the spiking domain.

The classification accuracy of the target model on reconstructed samples (\( A_T \)) reveals a stark contrast between ANNs and SNNs. In the ANN case, the target model classifies its own reconstructed samples with near-perfect accuracy (100\% on MNIST and 97.5\% on AT\&T Face), demonstrating that the generated samples align well with the target model’s learned decision boundaries. However, for SNNs, \( A_T \) is significantly lower (70\% for MNIST and 49.2\% for AT\&T Face), implying that the reconstructed images fail to retain essential class-discriminative features.
\( A_T \) also serves as an indicator of convergence, indirectly measuring how well the optimization problem in Eq.~\ref{eq:MI-opt} is addressed.

\begin{figure}[h]
  \centering
  \includegraphics[width=\columnwidth]{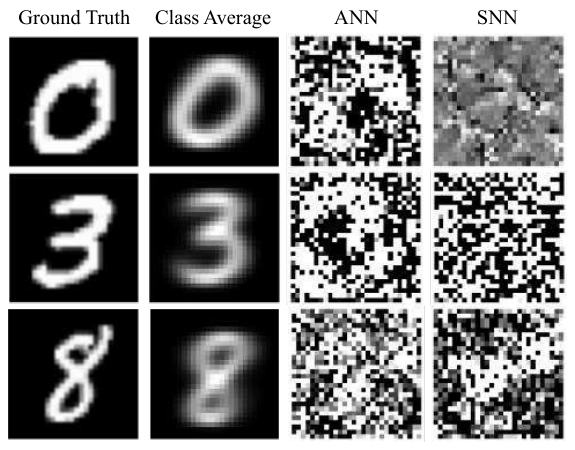}
  \caption{Qualitative assessment of reconstructed MNIST samples.}
  \label{fig:visual_mnist}
\end{figure}

\begin{figure}[ht]
  \centering
  \includegraphics[width=\columnwidth]{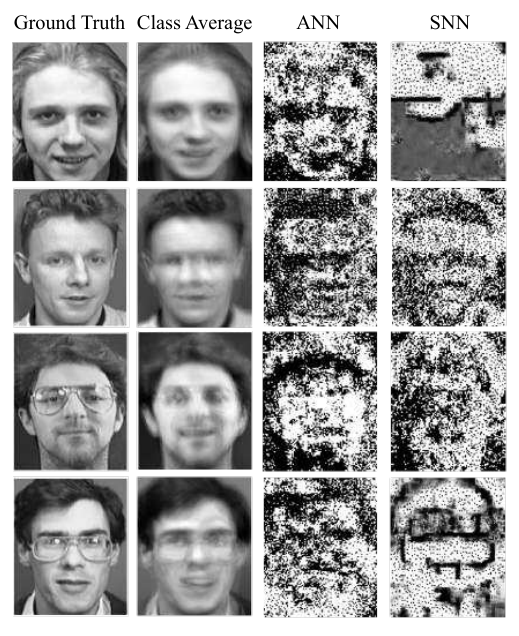}
  \caption{Qualitative assessment of reconstructed AT\&T Face dataset samples.}
  \label{fig:visual_face}
\end{figure}

This is further supported by the qualitative assessment of reconstructed images in Figures~\ref{fig:visual_mnist} and~\ref{fig:visual_face}, which display reconstructions for MNIST and AT\&T Face, respectively. While ANN-based reconstructions retain more discernible features of the original data, SNN-based reconstructions are significantly degraded, and in some cases, the attack fails to converge entirely, producing incoherent outputs, as seen in the reconstructed digit ‘0’ in Figure~\ref{fig:visual_mnist}.

A surprising countertrend emerges in surrogate fidelity (\( F_S \)), which is higher for SNNs than for ANNs, despite the significantly lower surrogate test accuracy (\( A_S \)). This suggests that while the surrogate model effectively mimics the SNN’s decision boundaries when queried with random inputs, it generalizes poorly to structured inputs such as the test set.
This discrepancy indicates that SNNs have fundamentally different decision boundary characteristics compared to ANNs. One plausible explanation is that SNNs encode information in a more discrete and temporally distributed manner, leading to more fragmented or irregular decision boundaries. As a result, the surrogate can closely approximate these boundaries using randomly sampled inputs (resulting in high \( F_S \)) but fails to interpolate effectively when exposed to structured data distributions (leading to low \( A_S \)).

Furthermore, the encoding and decoding mechanisms in SNNs may contribute to attack resistance. Unlike ANNs, which compute activations in a single forward pass, SNNs accumulate information over multiple time steps. This distributed processing could act as an implicit form of obfuscation, making it more challenging for the generator to align its outputs with the target model’s decision patterns.

Overall, these findings provide strong evidence that SNNs exhibit an inherent degree of robustness against black-box MI attacks.
Unlike ANNs, where the generator can exploit smooth decision boundaries to refine reconstructions, the event-driven computations in SNNs appear to introduce irregularities that degrade the inversion process. However, the high surrogate fidelity on random inputs raises interesting questions about the nature of SNN decision boundaries and whether alternative encoding schemes (e.g., latency coding instead of rate coding) influence SNN privacy. Additionally, extending this analysis to neuromorphic datasets with real temporal dependencies, such as IBM DvsGesture \cite{amir2017low}, would provide a broader understanding of SNN privacy in real-world applications.

By highlighting these distinctions, this work contributes to a deeper understanding of adversarial threats in neuromorphic computing and establishes a foundation for future research into privacy-preserving spiking architectures.

\section{Conclusion}
\label{sec:conclusion}

\noindent
In this work, we investigated the privacy characteristics of SNNs under black-box MI attacks, assessing whether their discrete and event-driven computations offer inherent resistance to adversarial data reconstruction. Using GAMIN, a black-box agnostic MI framework, we systematically evaluated the effectiveness of the attack on SNNs compared to traditional ANNs. Our findings reveal that SNNs exhibit stronger resistance to model inversion, with reconstructions displaying severe degradation, unstable attack convergence, and reduced effectiveness across multiple evaluation metrics.

A key factor contributing to this resistance appears to be the structure of the SNN decision boundaries, which differ fundamentally from those of ANNs. The discrete and temporally distributed nature of spike-based computations introduces irregular decision regions that disrupt the surrogate model’s ability to generalize, limiting the attack’s ability to approximate the target model effectively. Furthermore, the encoding and decoding mechanisms inherent to SNNs may introduce an additional layer of obfuscation, restricting the generator’s capacity to synthesize meaningful training samples.

Future research will investigate whether alternative spiking encodings, such as latency coding \cite{auge2021survey}, impact inversion resistance, and assess attack performance on neuromorphic datasets with real temporal dependencies, such as event-based DVS datasets. 

This study contributes to the understanding of privacy in neuromorphic computing, providing new insights into the adversarial risks associated with SNNs. As these architectures gain traction in privacy-sensitive applications, a deeper exploration of their privacy-preserving properties will be essential for the development of privacy-aware and resilient neuromorphic systems.

\section*{Acknowledgment}
The research was funded in part by National Science Foundation through award CCF2319619 and Department of Energy through award DE-SC0025349.

\nocite{*}
\bibliographystyle{IEEEtran}
\bibliography{ref}
\end{document}